\documentclass{article}
\usepackage{iclr2025_conference,times}

\usepackage{amsmath,amsfonts,bm}

\def\eqref#1{equation~\ref{#1}}

\def\1{\bm{1}}

\DeclareMathAlphabet{\mathsfit}{\encodingdefault}{\sfdefault}{m}{sl}
\SetMathAlphabet{\mathsfit}{bold}{\encodingdefault}{\sfdefault}{bx}{n}

\usepackage{hyperref}
\usepackage{url}
\usepackage{graphicx}
\usepackage{booktabs}
\usepackage{multirow}
\usepackage{wrapfig}

\title{Exploring Ordinal Bias in Action Recognition for Instructional Videos}

\author{Joochan Kim\thanks{Work was done when the author was a master's student at Seoul National University} \\
Korea Institute of Science and Technology \\
Republic of Korea \\
\texttt{joochan.k@kist.re.kr} \\
\And
Minjoon Jung \\
Seoul National University \\
Republic of Korea \\
\texttt{mjjung@bi.snu.ac.kr} \\
\And
Byoung-Tak Zhang\thanks{Corresponding Author} \\
Seoul National University \\
Republic of Korea \\
\texttt{btzhang@bi.snu.ac.kr} \\
}

\iclrfinalcopy

\begin{document}

\maketitle

\begin{abstract}
Action recognition models have achieved promising results in understanding instructional videos.
However, they often rely on dominant, dataset-specific action sequences rather than true video comprehension, a problem that we define as ordinal bias.
To address this issue, we propose two effective video manipulation methods: Action Masking, which masks frames of frequently co-occurring actions, and Sequence Shuffling, which randomizes the order of action segments.
Through comprehensive experiments, we demonstrate that current models exhibit significant performance drops when confronted with nonstandard action sequences, underscoring their vulnerability to ordinal bias.
Our findings emphasize the importance of rethinking evaluation strategies and developing models capable of generalizing beyond fixed action patterns in diverse instructional videos.
\end{abstract}

\section{Introduction}
\label{sec:intro}
\begin{wrapfigure}{r}{0.5\textwidth}
\centering
\vspace{-5mm}
\includegraphics[width=0.42\textwidth]{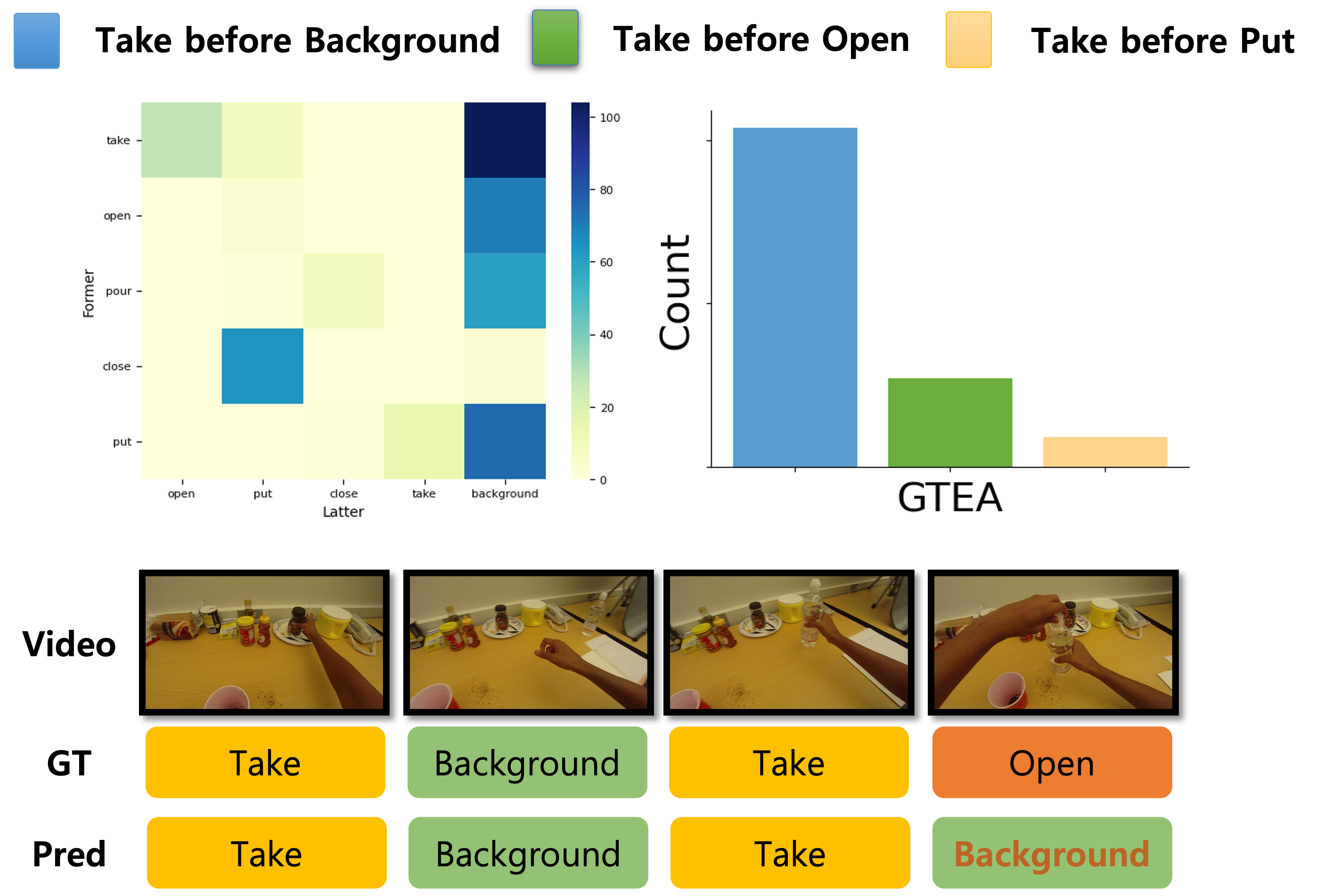}
\vspace{-2mm}
\caption{
\textbf{Illustration of the ordinal bias.} 
Due to the dominant action pair `Take-Background', the model fails to predict the action `Open.'
}
\label{fig:intro}
\end{wrapfigure}

Action recognition in instructional videos has witnessed remarkable progress, primarily driven by models that excel in curated benchmark datasets \citep{farha2019ms, ishikawa2021alleviating, li2020ms, yi2021asformer}.
However, these datasets often present a limited view of real-world variability by favoring specific, repeated action sequences.
In contrast, real-world instructional videos, such as those for cooking or physical exercise, frequently present more diverse action sequences than those captured in benchmark datasets.
As a result, models trained on conventional benchmarks tend to exploit these spurious ordinal patterns that we refer to as \textbf{\textit{ordinal bias}}.

We observe that existing datasets \citep{fathi2011learning, stein2013combining, kuehne2014language} demonstrate biased action sequences, which lead the model to suffer from spurious correlations.
As shown in \autoref{fig:intro}, the dataset exhibits a dominant occurrence of the action `Take' followed by `Background.'
This biases the model toward learning spurious correlations, causing it to predict `Background' as the next action rather than relying on visual inputs to correctly predict `Open.'
This raises concerns about the reliability of evaluations and the risk of overestimating the performance of the action recognition models.

To address this, we propose two video manipulation methods: \textbf{\textit{Action Masking}} and \textbf{\textit{Sequence Shuffling}} for a reliable evaluation.
In the action masking method, we selectively mask or replace the video frames corresponding to a specific action unit with a `no action' label, compelling the model to depend on alternative contextual visual cues rather than on learned ordinal patterns.
In contrast, the sequence shuffling method randomly rearranges the order of the action labels while keeping the frame order within each action unit intact.

With our methods, our experiments reveal that state-of-the-art action recognition models struggle to generalize manipulated videos, demonstrating their lack of robustness. Furthermore, even when models are trained on videos with mitigated action distributions through our manipulation techniques, they still tend to capture dominant action pairs in datasets. These findings highlight the pressing need to rethink an evaluation framework, a training strategy, and advanced modeling so that models can adapt to real-world scenarios.

Our contributions can be summarized as follows:
\begin{itemize}
    \item We study the ordinal bias problem in instructional video understanding, where action recognition models tend to capture dominant action patterns in datasets rather than achieving true video comprehension.
    \item To systematically address this issue, we propose two video manipulation techniques, \emph{Action Masking} and \emph{Sequence Shuffling}, which mitigate dataset bias and expose models to a broader range of action sequences.
    \item Our extensive experiments reveal significant deficiencies in existing models, underscoring the need for further research and improved evaluation frameworks.
\end{itemize}

\section{Related Work}
\label{sec:relatedwork}

\subsection{Recognition of Actions in Instructional Videos}
Instructional video analysis has emerged as a prominent area of research in the field of video comprehension.
In particular, multiple datasets of instructional videos \citep{fathi2011learning, stein2013combining, kuehne2014language} have been introduced, offering extensive contextual information on human activities.
Although various methods have been proposed \citep{farha2019ms, li2020ms, ishikawa2021alleviating, yi2021asformer, liu2023diffusion}, and \citet{li2022bridge} explores the understanding of ordinal action by explicitly modeling inter-action context, the ordinal bias issue can overestimate their performance.
This comes from their exploitation of the action sequence observed during training stage.

\subsection{Bias in Action Recognition}
Recent work has advanced reliable video systems and evaluations across various video understanding tasks.
\citet{otani2020uncovering, yuan2021closer, jung2024consistency} have analyzed the reliable evaluation of temporal understanding, and \citet{li2018resound, hara2021rethinking} have raised the bias problem in action recognition.
Consequently, a line of works \citep{soar2023, li2024fairaction} have explored the dual challenges of background and foreground biases, demonstrating that action recognition models can be inadvertently biased by static and dynamic cues.
Although the previous works \citep{nam2020learning, duan2022mitigating} have addressed the bias by retraining the model, these approaches are computationally expensive and overlook the underlying imbalance of the data set.
In contrast, we directly address the issue by manipulating the video data itself and provide insights into how current models perform if they are trained with these video variants.

\section{Experimental Setup}
In this section, we first introduce three video datasets and five action recognition models for our experiment. Then we provide details on the evaluation metrics.

\subsection{Dataset}
We utilize three action recognition datasets: Georgia Tech Egocentric Activities (GTEA) \citep{fathi2011learning}, 50Salads~\citep{stein2013combining}, and Breakfast~\citep{kuehne2014language}.
GTEA includes 28 videos depicting daily kitchen activities, featuring 11 action categories.
Each video has an average of 20 action units and a duration of approximately 30 seconds.
50Salads contains 50 videos of actors preparing salads in various kitchen environments, with more than 20 actors participating.
The videos in 50Salads are more than six minutes long and cover 17 action categories.
Breakfast consists of over 1700 videos that contain breakfast preparation scenes and has 48 action categories.
This dataset has the most complex labeling scheme among the three datasets.

\subsection{Action Recognition Models}
We consider five distinct models: MS-TCN \citep{farha2019ms}, MS-TCN++ \citep{li2020ms}, ASRF \citep{ishikawa2021alleviating}, and DiffAct \citep{liu2023diffusion}.
MS-TCN employs a multi‑stage architecture with dilated temporal convolutions and a smoothing loss to iteratively refine frame‑level predictions.
MS-TCN++ extends this approach by integrating dual dilated layers that capture both local and global contexts while decoupling prediction generation from refinement.
ASRF improves segmentation quality by adding an auxiliary branch that explicitly regresses action boundaries to mitigate over‑segmentation errors.
ASFormer leverages a Transformer‑based framework augmented with temporal convolutions and a hierarchical representation pattern for iterative prediction refinement.
Lastly, DiffAct formulates action segmentation as a conditional sequence generation task that iteratively denoises a noisy action sequence by leveraging priors such as positional, boundary, and relational cues.
For our experiments, we utilized I3D \citep{carreira2017quo} video features, pre-trained on the Kinetics dataset \citep{kay2017kinetics}, and a single NVIDIA RTX 3090.

\subsection{Evaluation Metric}
To evaluate the performance of the model, we adopt frame-wise accuracy, a primary metric that gauges the percentage of accurately classified actions within a unit frame of a test dataset.
To produce a result, we use 5-fold cross-validation to evaluate the proposed approach's performance on the 50Salads dataset.
For the remaining datasets, 4-fold cross-validation is performed to estimate the average performance measure.

\begin{figure}[t]
\begin{center}
\includegraphics[width=0.9\textwidth]{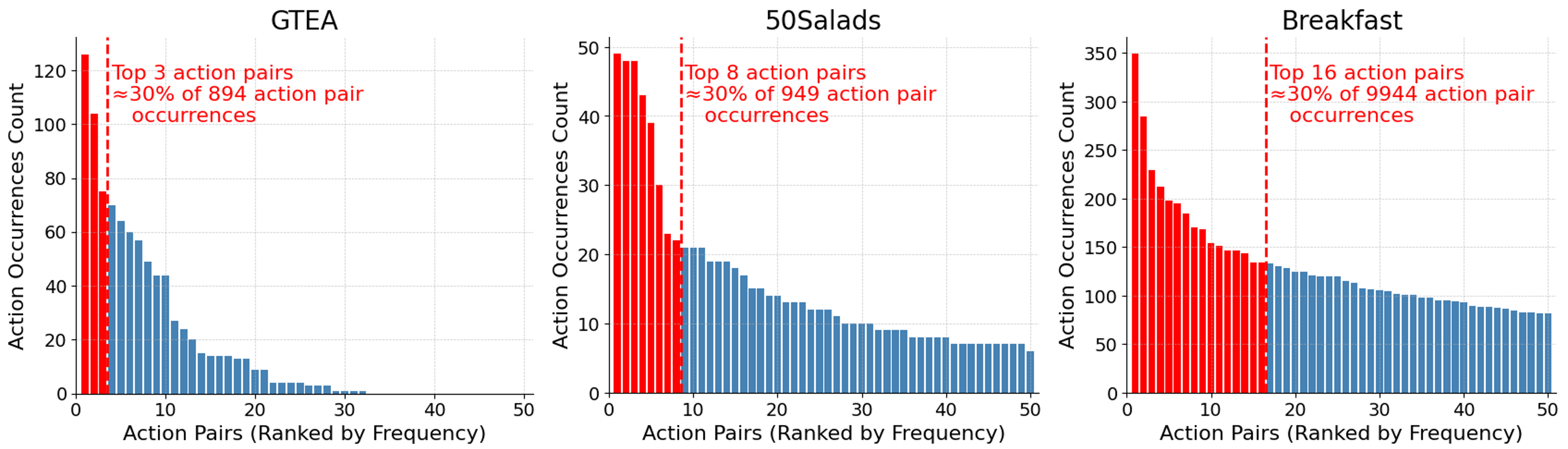}
\caption{
\textbf{Long-tailed distributions of action pairs in datasets.}
Each dataset's histogram represents the frequency of action pairs, ranked by their occurrence count.
The red-highlighted action pairs contribute to 30\% of all actions in the dataset, despite being a small fraction of the total pairs.
We only report up to top 50 action pairs in this visualization.
}
\label{fig:dist}
\end{center}
\end{figure}

\section{Ordinal Bias Problem}
\label{sec:problem}

\subsection{Long tail distribution of action pairs}
We begin by analyzing the distribution of action pairs across the datasets.
As shown in \autoref{fig:dist}, each dataset exhibits a pronounced long-tailed pattern.
In detail, in the Breakfast dataset, 16 of 228 action pairs represent 30\% of all action pair occurrences.
Similarly, in the 50Salads dataset, 8 of 120 action pairs contribute to 30\% of the total occurrences of action pairs", and in the GTEA dataset, only 3 of 32 action pairs comprise 30\% of the total action pair occurrences.
This skewed distribution may lead to biased predictions, as models can become overly influenced by the few frequently occurring action pairs, potentially misrepresenting the diversity of real-world instructional videos.
To address this issue and enable more reliable evaluations, we introduce video manipulation methods designed to counteract the effects of this long-tailed distribution.

\begin{figure}[t]
\begin{center}
\includegraphics[width=0.9\textwidth]{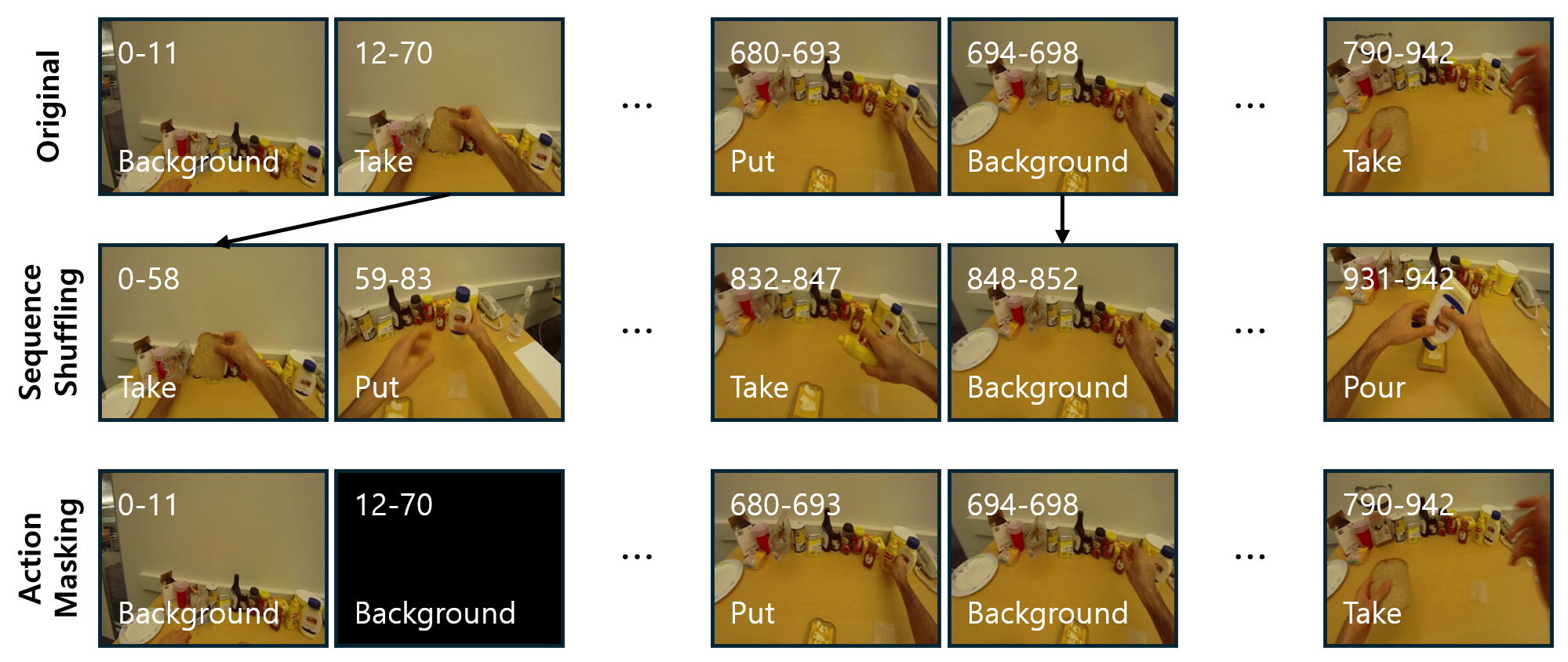}
\caption{
\textbf{Manipulation techniques}.
Each video contains 943 frames. A single image represents consecutive frames, displayed in the top-left corner, while the action label is in the bottom-left corner.
In the sequence shuffling, frames are shuffled in sequences, e.g., moving frames 12 to 70 to positions 0 to 58, and frames 695 to 698 to positions 848 to 852.
In action masking, frames 12 to 70 are masked and labeled as background.
}
\label{fig:video}
\end{center}
\end{figure}

\subsection{Video Manipulation Methodologies}
\label{subsec:tech}
We propose two video manipulation techniques, Action Masking and Sequence Shuffling, as shown in \autoref{fig:video}.
For action masking, we mask the video frames of a specific action unit, and the corresponding action label is replaced with `no action.'
By doing so, we verify whether the model predicts `no action' accounting for visual variants or if it makes biased predictions.
The second method, sequence shuffling, randomly rearranges the order of action segments without altering the frame order within each action unit.
This allows us to construct datasets with distinctive label distributions, mitigating the influence of skewed distributions, and enabling reliable evaluations.
Importantly, sequence shuffling preserves the internal semantic coherence of each action unit since the visual continuity of each segment remains intact.
Although altering global action sequences may challenge models that implicitly rely on positional cues, this manipulation explicitly targets the assessment of a model's robustness against ordinal bias.
Consequently, sequence shuffling facilitates rigorous evaluation of model performance in scenarios involving unfamiliar action orders, ensuring meaningful insights into their generalization capability.
Further study can be found in Appendix \ref{sec:app_manipulation}.

\begin{figure*}[t]
\begin{center}
\includegraphics[width=0.9\textwidth]{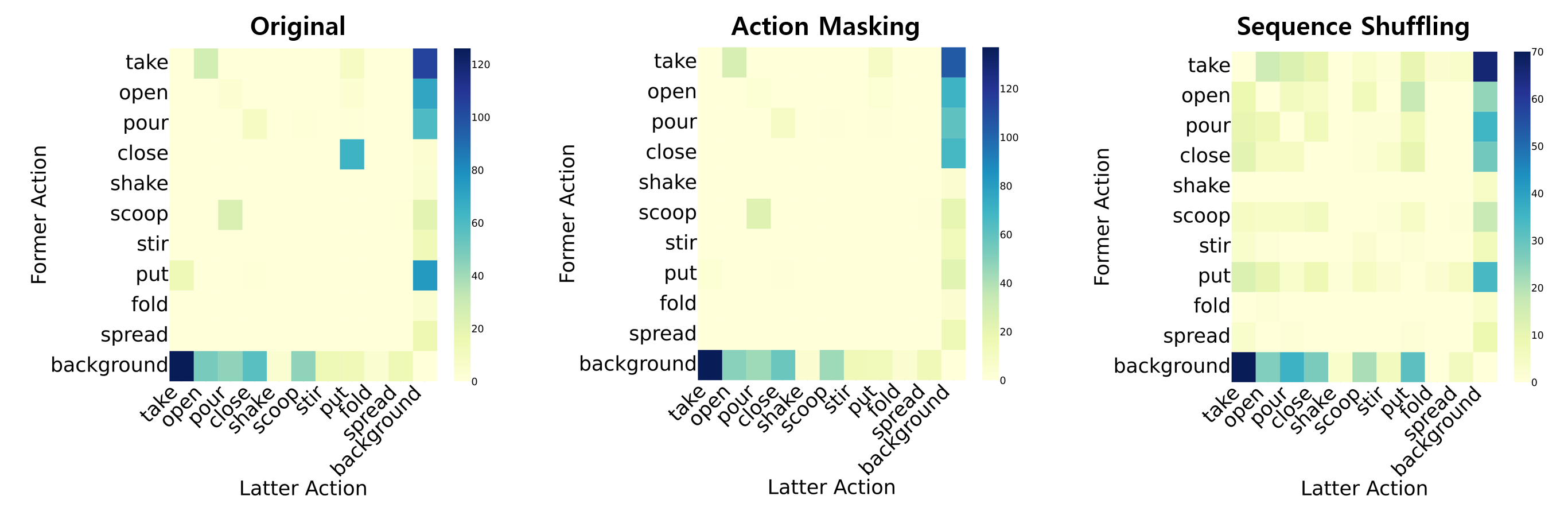}
\caption{
\textbf{Heatmap of the frequency of action pairs with GTEA dataset}.
We use the initial action `close' for action masking.
}
\label{fig:heat_gtea}
\end{center}
\end{figure*}

\subsection{Evaluation with Proposed Manipulation Methods}
\autoref{fig:heat_gtea} shows the result of our methods, demonstrating that our manipulation methods successfully change the distributions of action pairs.
Specifically, for action masking, all the subsequent labels of `close' have been switched to `background', whereas the sequence shuffling reduced the maximum value of high occurrences and introduced a new pair of actions.
Also, in the sequence shuffling, the number of existing biased pairs is decreased while previously absent action pairs are created.
More visual examples can be found in Appendix \ref{subsec:app_heat}

Now, we apply the action masking technique to the original dataset and conduct an experiment to see how the model trained with the original dataset behaves when it encounters a masked section.
We first select the action pair that is frequently seen in the original dataset.
We selected action pairs according to these criteria:
1) Considering the initial action, we observe the frequency of subsequent actions to determine if a particular combination is significantly more common compared to the others.
2) Subsequent actions should not equate to `no action.'
As a result, in the GTEA dataset, `close' is used as a prior action, making up about 7.5\% of the entire dataset, with the `put' combination comprising approximately 95.5\% of these actions.
Similarly, in the Breakfast dataset, `pour\_dough2pan' serves as the initial action, accounting for around 1.6\% of the total dataset, while `fry\ pancake' constitutes nearly 91.1\% of the subsequent actions.
Then, we mask frames that correspond to the latter action unit and replace its action label with `no action.'
Lastly, we make the model predict the masked parts and inspect the accuracy.

\begin{figure}[t]
\begin{center}
\includegraphics[width=0.9\textwidth]{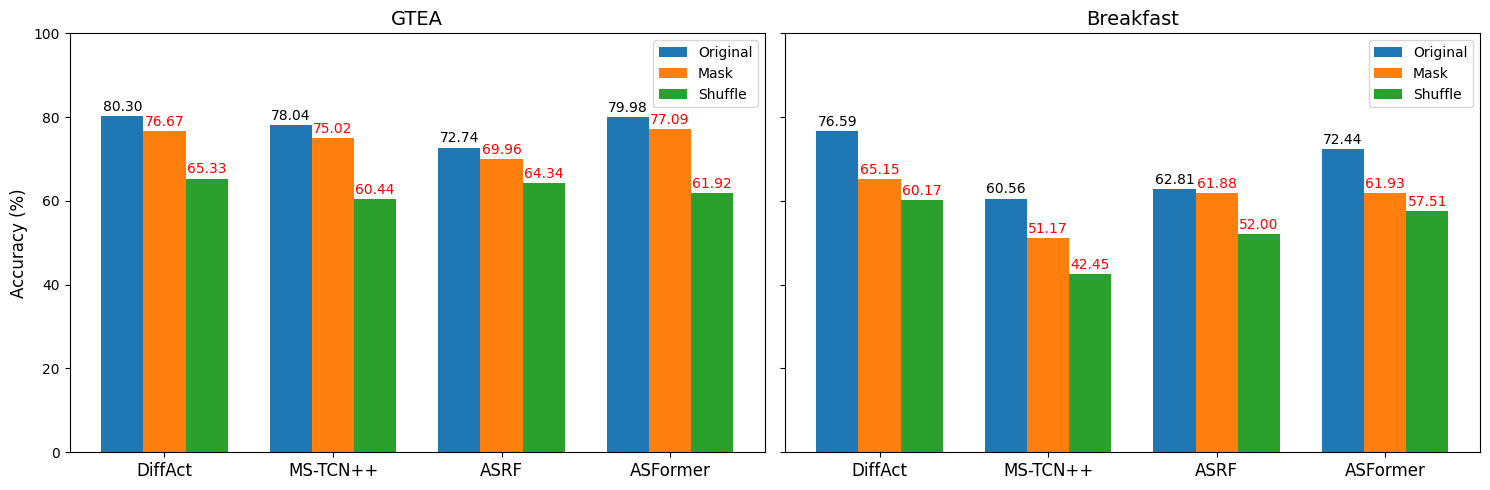}
\caption{\textbf{The results of model tested on the original and manipulated set}. Consistent performance drops across different datasets and models suggest that current models rely on the ordinal bias in the datasets.
}
\label{fig:og_mp}
\end{center}
\end{figure}

\autoref{fig:og_mp} shows the results of our experiment, demonstrating that the model finds it difficult to accurately predict the manipulated test videos.
This result indicates that the model misclassifies masked regions as having an action label from the original dataset instead of identifying them as `no action,' suffering from the ordinal bias problem.
The result also reveals that the model does not utilize visual information, but exploits spurious correlation for prediction.
We will discuss the ordinal bias problem in detail in the next section.

\section{Analysis of Ordinal Bias Problem}
\label{sec:analysis}
Within this section, we explore how models contribute to the issue of ordinal bias.

\subsection{Evaluation of Model Generalization}
\label{subsec:general}
We investigate the extent to which a model is responsible for the problem by training it on a manipulated dataset using the sequence shuffling method described in Section \ref{subsec:tech}.
Here, we skip the action masking method as it only introduces no-action labels; therefore, specific actions are only followed by `no-action.'
In contrast, sequence shuffling provides more diverse action pairs, allowing us to assess how models handle varied action patterns.
We then evaluate the performance of the model on the original dataset and sequence shuffling.
A model with satisfactory generalization should exhibit good performance in the original dataset despite being trained on a manipulated one.
Table \ref{tab:model} shows a significant performance discrepancy between the model trained on the modified dataset (S/O) and the superior performance of the model trained on the unchanged dataset (O/O).

\begin{figure}[t]
\begin{center}
\includegraphics[width=0.9\textwidth]{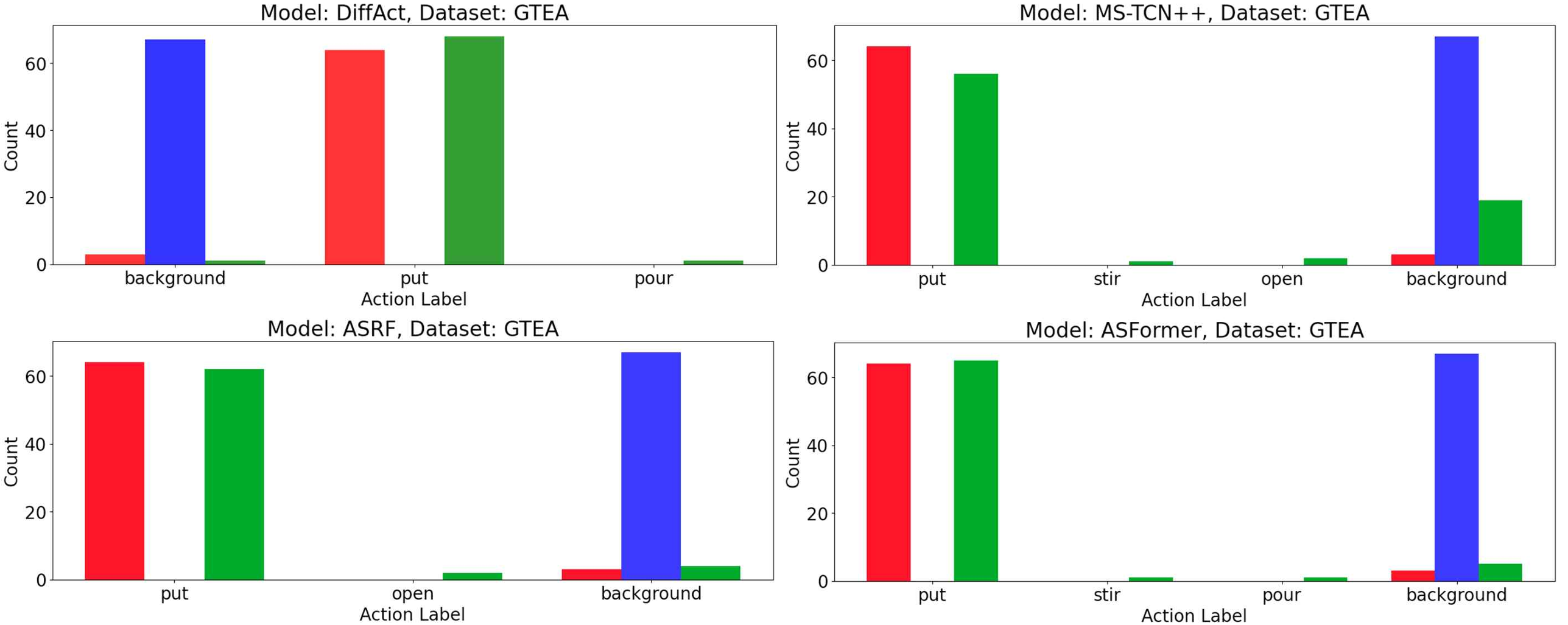}
\caption{
\textbf{Distribution of the action labels on GTEA dataset}.
\textcolor{red}{Red}: distribution of original dataset label;
\textcolor{blue}{Blue}: distribution of action masking dataset label;
\textcolor{green}{Green}: distribution of model predicted label.
For more visualization, please refer to the Appendix \ref{subsec:qual_mask} and \ref{subsec:qual_shuf}.
}
\label{fig:mask}
\end{center}
\end{figure}

Furthermore, we have investigated whether these results come from ordinal bias by comparing the label distribution among the original dataset, the manipulated dataset, and the model's predictions.
If the model is robust, its prediction distribution (green) should resemble the manipulated dataset distribution (blue), rather than the original dataset distribution (red).
\autoref{fig:mask} shows results, which implies that the model tends to make predictions by following the trend of the training dataset, not by utilizing given visual information.
This outcome implies that the model exploits spurious correlations during inference to achieve higher scores, resulting in an overestimation.
Therefore, a model must have an improved generalization capability to reduce the ordinal bias.

\begin{table}[t]
\begin{center}
\resizebox{0.9\textwidth}{!}{
\begin{tabular}{lccccccccccccccc}
\toprule
\multirow{2}{*}{Dataset} & \multicolumn{3}{c}{MS-TCN} & \multicolumn{3}{c}{MS-TCN++} & \multicolumn{3}{c}{ASRF} & \multicolumn{3}{c}{ASFormer} & \multicolumn{3}{c}{DiffAct} \\ 
\cmidrule(l){2-4} \cmidrule(l){5-7} \cmidrule(l){8-10} \cmidrule(l){11-13} \cmidrule(l){14-16}  
 & O/O & S/O & S/S & O/O & S/O & S/S & O/O & S/O & S/S & O/O & S/O & S/S & O/O & S/O & S/S \\ \midrule
GTEA      & 76.12  & 64.50  & 69.63  & 78.04  & 65.08  & 67.52  & 72.74  & 58.08  & 69.46  & 79.98  & 71.61  & 76.80  & 80.30  & 72.68  & 76.89  \\ 
Breakfast & 66.98  & 50.08  & 56.13  & 60.56  & 47.96  & 54.77  & 62.81  & 54.27  & 57.30  & 72.44  & --     & --     & 76.59  & --     & --     \\ 
50Salads  & 79.33  & 69.85  & 71.17  & 74.89  & 72.54  & 73.43  & 82.14  & 65.13  & 70.69  & 85.62  & 65.41  & 66.93  & 88.43  & 76.20  & 78.39  \\ 
\bottomrule
\end{tabular}
}
\caption{
\textbf{Accuracy of the models trained on the sequence shuffling dataset.}
\textbf{O/O}: model trained and tested on the original dataset;
\textbf{S/O}: model trained on the sequence shuffling dataset and tested on the original test set;
\textbf{S/S}:  model trained on the sequence shuffling dataset and tested on its test set.
Due to memory constraints, we were not able to test DiffAct and ASFormer on Breakfast.
}
\label{tab:model}
\end{center}
\end{table}

\subsection{Impact of Additional Training}
\label{sec:train}
\begin{wraptable}{r}{0.5\textwidth}
\vspace{-4mm}
\centering
\resizebox{0.48\textwidth}{!}{
\begin{tabular}{l cccc cccc}
\toprule
\multirow{2}{*}{Dataset} & \multicolumn{4}{c}{MS-TCN++} & \multicolumn{4}{c}{ASFormer} \\
\cmidrule(lr){2-5} \cmidrule(lr){6-9}
 & O/O & C/O & C/S & C/M & O/O & C/O & C/S & C/M \\
\midrule
GTEA      & 78.04 & 70.28 & 69.20 & 75.77 & 79.98 & 76.91 & 72.80 & 77.91    \\
Breakfast & 60.56 & 50.75 & 49.41 & 46.42 & 72.44 & -     & -     & -    \\
\bottomrule
\end{tabular}
}
\caption{
\textbf{Accuracy of the models trained with additional datasets.}
\textbf{O/O}: Trained and tested on the original dataset;
\textbf{C/O}: Trained on the combined dataset, tested on the original;
\textbf{C/S}: Trained on the combined, tested on the sequence shuffling dataset;
\textbf{C/M}: Trained on the combined, tested on the action masking dataset.
}
\label{tab:finetune}
\end{wraptable}

In many bias-related problems, the incorporation of additional data helps alleviate bias.
Therefore, we investigate whether training models with an additional augmented dataset can mitigate ordinal bias in action recognition.
To this end, we designed a curriculum learning-like strategy by sequentially training the model on three variants of the dataset: the original, the masked, and the shuffled versions.
This progression is intended to gradually expose the model to increasing difficulty levels and reduce its reliance on spurious correlations.
However, as highlighted in Section \ref{subsec:general}, the use of action masking is not suitable for model training.
To overcome this, we used a strategy for action masking inspired by Masked Language Modeling \citep{devlin2018bert}, where actions are randomly hidden with a likelihood of 15\%, instead of our original approach.
In contrast, for sequence shuffling, we utilized the method we had proposed.
Consequently, every augmented dataset retains its original size, allowing the model to be trained on a dataset that is triple the size of the initial one, which we refer to as the `Combined' dataset.

However, as shown in \autoref{tab:finetune}, the model trained with an additional dataset did not exceed the performance of the model trained solely on the original dataset.
These results indicate that simply augmenting the training data, even through a curriculum-learning-like approach, does not effectively mitigate ordinal bias. This suggests that the bias is deeply ingrained in the training dynamics, and additional intervention, such as architectural modifications or specialized loss functions, may be required to address the issue.
The ablation study can be found in the Appendix \ref{sec:app_ab}.

\section{Conclusion}
Our investigation of ordinal bias reveals a critical oversight in current action recognition research: the overreliance on fixed, dataset-specific action sequences.
Although high accuracies are reported on popular benchmarks, such performance does not necessarily translate into reliable predictions in real-world settings, where the sequence of actions is highly variable and unpredictable.
By applying our proposed video manipulation techniques, we demonstrate that models vulnerable to ordinal bias exhibit significant drops in performance when faced with non-standard action orders.

Regardless, future work could continue to explore:
(1) developing more robust action recognition models that effectively generalize beyond fixed action sequences,
(2) organizing instructional video datasets that are not affected by dominant action pairs, and
(3) constructing a more automatic approach to identify the existence of ordinal biases.
Addressing these directions will be crucial to achieving reliable and consistent action recognition in various instructional video scenarios.

\section*{Acknowledgement}
This research is supported by the IITP (RS-2021-II212068-AIHub/10\%, RS-2021-II211343-GSAI/15\%, RS-2022-II220951-LBA/15\%, RS-2022-II220953-PICA/20\%), NRF (RS-2024-00353991-SPARC/20\%, RS-2023-00274280-HEI/10\%), and KEIT (RS-2024-00423940/10\%) grant funded by the Korean government.

\bibliography{iclr2025_conference}

@inproceedings{fathi2011learning,
  title={Learning to recognize objects in egocentric activities},
  author={Fathi, Alireza and Ren, Xiaofeng and Rehg, James M},
  booktitle={CVPR 2011},
  pages={3281--3288},
  year={2011},
  organization={IEEE}
}

@inproceedings{stein2013combining,
  title={Combining embedded accelerometers with computer vision for recognizing food preparation activities},
  author={Stein, Sebastian and McKenna, Stephen J},
  booktitle={Proceedings of the 2013 ACM international joint conference on Pervasive and ubiquitous computing},
  pages={729--738},
  year={2013}
}

@inproceedings{kuehne2014language,
  title={The language of actions: Recovering the syntax and semantics of goal-directed human activities},
  author={Kuehne, Hilde and Arslan, Ali and Serre, Thomas},
  booktitle={Proceedings of the IEEE conference on computer vision and pattern recognition},
  pages={780--787},
  year={2014}
}

@article{otani2020uncovering,
  title={Uncovering hidden challenges in query-based video moment retrieval},
  author={Otani, Mayu and Nakashima, Yuta and Rahtu, Esa and Heikkil{\"a}, Janne},
  journal={arXiv preprint arXiv:2009.00325},
  year={2020}
}

@inproceedings{yuan2021closer,
  title={A closer look at temporal sentence grounding in videos: Dataset and metric},
  author={Yuan, Yitian and Lan, Xiaohan and Wang, Xin and Chen, Long and Wang, Zhi and Zhu, Wenwu},
  booktitle={Proceedings of the 2nd international workshop on human-centric multimedia analysis},
  pages={13--21},
  year={2021}
}

@article{jung2024consistency,
  title={On the Consistency of Video Large Language Models in Temporal Comprehension},
  author={Jung, Minjoon and Xiao, Junbin and Zhang, Byoung-Tak and Yao, Angela},
  journal={arXiv preprint arXiv:2411.12951},
  year={2024}
}

@article{yi2021asformer,
  title={Asformer: Transformer for action segmentation},
  author={Yi, Fangqiu and Wen, Hongyu and Jiang, Tingting},
  journal={arXiv preprint arXiv:2110.08568},
  year={2021}
}

@inproceedings{ishikawa2021alleviating,
  title={Alleviating over-segmentation errors by detecting action boundaries},
  author={Ishikawa, Yuchi and Kasai, Seito and Aoki, Yoshimitsu and Kataoka, Hirokatsu},
  booktitle={Proceedings of the IEEE/CVF winter conference on applications of computer vision},
  pages={2322--2331},
  year={2021}
}

@inproceedings{farha2019ms,
  title={Ms-tcn: Multi-stage temporal convolutional network for action segmentation},
  author={Farha, Yazan Abu and Gall, Jurgen},
  booktitle={Proceedings of the IEEE Conference on Computer Vision and Pattern Recognition},
  pages={3575--3584},
  year={2019}
}

@article{li2020ms,
   author={Shi-Jie Li and Yazan AbuFarha and Yun Liu and Ming-Ming Cheng and Juergen Gall},
    journal={IEEE Transactions on Pattern Analysis and Machine Intelligence}, 
    title={MS-TCN++: Multi-Stage Temporal Convolutional Network for Action Segmentation}, 
    year={2020},
    volume={},
    number={},
    pages={1-1},
    doi={10.1109/TPAMI.2020.3021756},
}

@inproceedings{carreira2017quo,
  title={Quo vadis, action recognition? a new model and the kinetics dataset},
  author={Carreira, Joao and Zisserman, Andrew},
  booktitle={proceedings of the IEEE Conference on Computer Vision and Pattern Recognition},
  pages={6299--6308},
  year={2017}
}

@inproceedings{li2018resound,
  title={Resound: Towards action recognition without representation bias},
  author={Li, Yingwei and Li, Yi and Vasconcelos, Nuno},
  booktitle={Proceedings of the European Conference on Computer Vision (ECCV)},
  pages={513--528},
  year={2018}
}

@inproceedings{li2022bridge,
  title={Bridge-prompt: Towards ordinal action understanding in instructional videos},
  author={Li, Muheng and Chen, Lei and Duan, Yueqi and Hu, Zhilan and Feng, Jianjiang and Zhou, Jie and Lu, Jiwen},
  booktitle={Proceedings of the IEEE/CVF conference on computer vision and pattern recognition},
  pages={19880--19889},
  year={2022}
}

@inproceedings{hara2021rethinking,
  title={Rethinking training data for mitigating representation biases in action recognition},
  author={Hara, Kensho and Ishikawa, Yuchi and Kataoka, Hirokatsu},
  booktitle={Proceedings of the IEEE/CVF Conference on Computer Vision and Pattern Recognition},
  pages={3349--3353},
  year={2021}
}

@article{nam2020learning,
  title={Learning from failure: De-biasing classifier from biased classifier},
  author={Nam, Junhyun and Cha, Hyuntak and Ahn, Sungsoo and Lee, Jaeho and Shin, Jinwoo},
  journal={Advances in Neural Information Processing Systems},
  volume={33},
  pages={20673--20684},
  year={2020}
}

@inproceedings{soar2023,
  title={SOAR: Scene-debiasing Open-set Action Recognition},
  author={Zhai, Yuanhao and Liu, Ziyi and Wu, Zhenyu and others},
  booktitle={arXiv preprint arXiv:2309.01265},
  year={2023}
}

@inproceedings{duan2022mitigating,
  title={Mitigating representation bias in action recognition: Algorithms and benchmarks},
  author={Duan, Haodong and Zhao, Yue and Chen, Kai and Xiong, Yuanjun and Lin, Dahua},
  booktitle={European Conference on Computer Vision},
  pages={557--575},
  year={2022},
  organization={Springer}
}

@article{li2024fairaction,
  title={Fair Action Recognition: Balancing Accuracy and Equity in Spatiotemporal Models},
  author={Li, Jian and others},
  journal={IEEE Transactions on Pattern Analysis and Machine Intelligence},
  year={2024},
  note={In press}
}

@inproceedings{liu2023diffusion,
  title={Diffusion action segmentation},
  author={Liu, Daochang and Li, Qiyue and Dinh, Anh-Dung and Jiang, Tingting and Shah, Mubarak and Xu, Chang},
  booktitle={Proceedings of the IEEE/CVF International Conference on Computer Vision},
  pages={10139--10149},
  year={2023}
}

@article{kay2017kinetics,
  title={The kinetics human action video dataset},
  author={Kay, Will and Carreira, Joao and Simonyan, Karen and Zhang, Brian and Hillier, Chloe and Vijayanarasimhan, Sudheendra and Viola, Fabio and Green, Tim and Back, Trevor and Natsev, Paul and others},
  journal={arXiv preprint arXiv:1705.06950},
  year={2017}
}

@article{devlin2018bert,
  title={Bert: Pre-training of deep bidirectional transformers for language understanding},
  author={Devlin, Jacob},
  journal={arXiv preprint arXiv:1810.04805},
  year={2018}
}
\bibliographystyle{iclr2025_conference}

\appendix

\renewcommand{\thefigure}{A\arabic{figure}}
\setcounter{figure}{0}
\renewcommand{\thetable}{A\arabic{table}}
\setcounter{table}{0}

\section{Evaluation of Action Segmentation Performance}
This section evaluates the capacity of the model to carry out the task of action segmentation.
Our model demonstrates proficiency in both action segmentation and recognition tasks.
We conduct an experiment on how altering instructional videos affects the model's performance in the action segmentation task by using the Shuffle method.
We begin by hypothesizing that the model's effectiveness on altered videos will be comparable to or exceed that on the original videos, due to the distinctly unnatural nature of action transitions.

However, the results presented in \autoref{tab:ab3} indicate that the performance of the ASFormer model on the manipulated video is inferior to that of the original video.
These outcomes suggest that the action segmentation task may also be influenced by ordinal bias, a matter we leave it as a futurework.

\begin{table}[ht]
\begin{center}
\begin{tabular}{@{}lcccccc@{}}
\toprule
\multirow{2}{*}{Metric} & \multicolumn{2}{c}{GTEA} & \multicolumn{2}{c}{Breakfast} & \multicolumn{2}{c}{50Salads} \\ \cmidrule(l){2-3} \cmidrule(l){4-5} \cmidrule(l){6-7}   
                        & O/O          & O/S         & O/O            & O/S            & O/O            & O/S           \\ \midrule
$Edit$                  & 84.04       & 50.71      & 54.01         & 39.09         & 73.50         & 37.17        \\
$F1@10$                 & 88.69       & 59.52      & 61.09         & 48.65         & 74.03         & 40.49        \\
$F1@25$                 & 87.76       & 57.51      & 58.15         & 46.09         & 68.69         & 36.63        \\
$F1@50$                 & 79.02       & 46.27      & 51.33         & 40.05         & 55.02         & 27.31        \\ \bottomrule
\end{tabular}
\caption{
\textbf{Performance of ASFormer model on the action segmentation task for various datasets.}
\textbf{O/O}: performance on original datasets. \textbf{O/S}: performance on action-shuffled dataset.
Metrics include segmental edit score and segmental overlap $F1$ score at a threshold of $k/100$ where $k$ equals the percentage of overlap, denoted as $F@k$.}
\label{tab:ab3}
\end{center}
\end{table}

\section{Ablation Experiment of Additional Training}
\label{sec:app_ab}
This section reports on the ablation study in an additional dataset.
As shown in \autoref{tab:ablation}, each outcome falls short of initial performance, indicating that further training may not address the ordinal bias issue.
Note that, as 50Salads does not have a label that refers to `no-action', we omit the results that use the action masking method.
Also, we have not presented the results for the Breakfast dataset when using ASFormer and DiffAct due to an inability to replicate these results.

\begin{table}[ht]
\centering
\begin{tabular}{c|l|cc|cc|cc}
\toprule
\multirow{2}{*}{Data} & \multirow{2}{*}{Method} 
& \multicolumn{2}{c}{GTEA} 
& \multicolumn{2}{c}{Breakfast}
& \multicolumn{2}{c}{50Salads} \\
& & O/O & C/O & O/O & C/O & O/O & C/O \\
\midrule
M+S & DiffAct   & 80.30 & 78.86 & -     & -     & -     & - \\
             & MS-TCN++  & 78.04 & 72.79 & 60.56 & 54.69 & -     & - \\
             & ASFormer  & 79.98 & 73.64 & -     & -     & -     & - \\
\midrule
O+M & DiffAct   & 80.30 & 78.00 & -     & -     & -     & - \\
             & MS-TCN++  & 78.04 & 75.39 & 60.56 & 59.73 & -     & - \\
             & ASFormer  & 79.98 & 79.39 & -     & -     & -     & - \\
\midrule
O+S & DiffAct   & 80.30 & 74.80 & -     & -     & 88.43 & 65.43 \\
             & MS-TCN++  & 78.04 & 69.78 & 60.56 & 54.85 & 74.89 & 72.79 \\
             & ASFormer  & 79.98 & 76.61 & -     & -     & 85.61 & 77.43 \\
\bottomrule
\end{tabular}
\caption{\textbf{Result of ablation study on additional dataset}
\textbf{O}: Original; \textbf{S}: Sequence Shuffling; \textbf{M}: Action Masking;
\textbf{O/O}: Model trained and tested on original dataset; \textbf{C/O}: Model trained on combined dataset and tested on original dataset.
}
\label{tab:ablation}
\end{table}

\section{Revisiting Video Manipulation Method}
\label{sec:app_manipulation}
Proposed manipulation technique is effective in judging whether the model uses visual cues well or not.
However, this methodology could lead to the following problems:
For the action masking method, the masked part of the frame may represent inaccurate inferences because it may be parts of the frame that the model did not encounter during training.
Also, sequence shuffling produces quite an unnatural video context, as we randomly shuffle sequences of actions.

To complement this issue, we use the sequence shuffling technique, but instead of shuffling randomly, we replace the latter action in frequently occurring action pairs with a random action and location within the video.
This technique will henceforth be referred to as `Limited Shuffling.'
\autoref{tab:var} shows experiment results, still revealing that the model suffers from an ordinal bias problem.
For qualitative results, refer to Appendix \ref{subsec:qual_shuf}.

\begin{table}[ht]
\centering
\begin{tabular}{lcccccccc}
\toprule
 & \multicolumn{2}{c}{DiffAct} & \multicolumn{2}{c}{MS-TCN++} & \multicolumn{2}{c}{ASRF} & \multicolumn{2}{c}{ASFormer} \\ \cmidrule(lr){2-3} \cmidrule(lr){4-5} \cmidrule(lr){6-7} \cmidrule(lr){8-9}
 & O/O & O/L & O/O & O/L & O/O & O/L & O/O & O/L \\
\midrule
GTEA      & 80.30 & 71.94 & 78.04 & 70.97 & 72.74 & 67.20 & 79.98 & 71.90 \\
Breakfast & 76.59 & 74.32 & 60.56 & 58.98 & 62.81 & 60.66 & 72.44 & 70.44 \\
50Salads  & 88.43 & 82.56 & 74.89 & 69.37 & 82.14 & 74.16 & 85.61 & 70.07 \\
\bottomrule
\end{tabular}
\caption{\textbf{Accuracy of model evaluated on Limited Shuffling method.}
\textbf{O/O}: performance on the original dataset
\textbf{O/L}: performance on limited shuffling method.
}
\label{tab:var}
\end{table}

\section{Detailed Qualitative Results}
\label{app:qualitative}
This section displays the visualization results mentioned in the main paper.

\subsection{Visualization of Action Pair Distribution}
\label{subsec:app_heat}
\begin{figure}[ht]
\begin{center}
\includegraphics[width=0.9\textwidth]{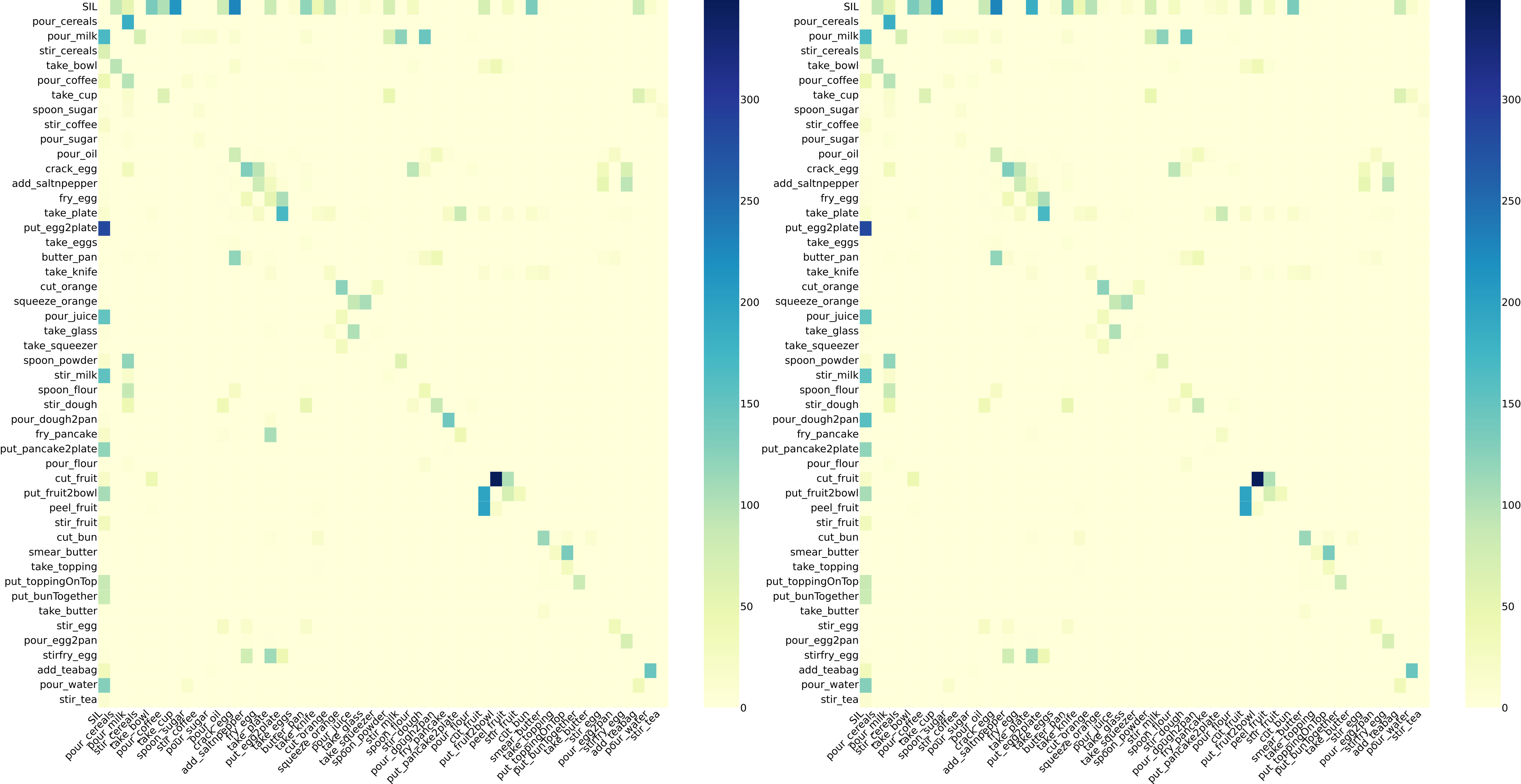}
\caption{
\textbf{Heatmap of the frequency of 2-gram action label pairs in Breakfast.}
The left is the original dataset and the right is the dataset with the action masking technique.
We use initial action as`pour\_dough2pan.'
The former action is represented on the Y-axis and the latter on the X-axis.
}
\label{fig:app_heat_mask}
\end{center}
\end{figure}

\begin{figure}[ht]
\begin{center}
\includegraphics[width=0.9\textwidth]{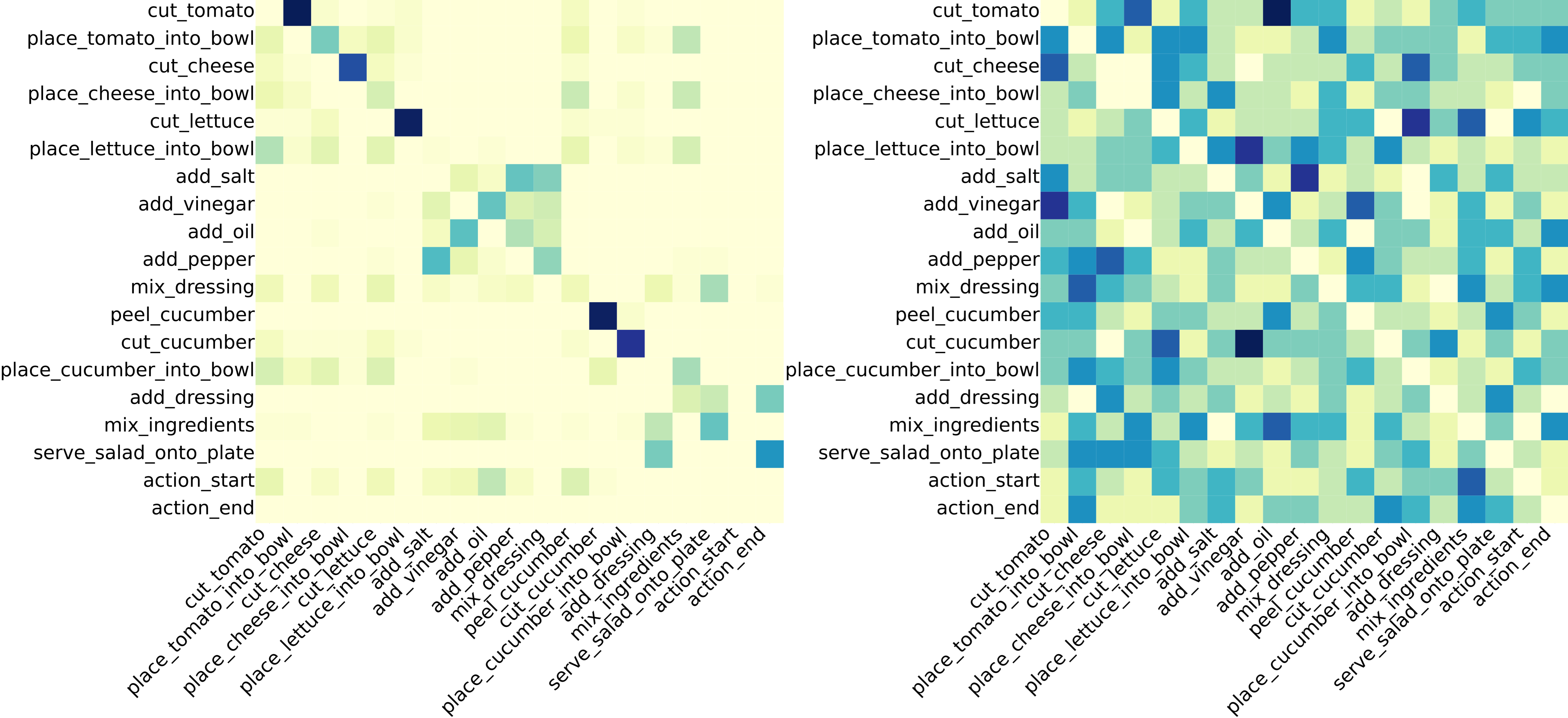}
\caption{
\textbf{Heatmap of the frequency of 2-gram action label pairs in the 50Salads dataset.}
The left displays the original dataset and the right shows the dataset with the Shuffle technique.
The former action is represented on the Y-axis and the latter action on the X-axis.
}
\label{fig:app_heat_50salads}
\end{center}
\end{figure}

\begin{figure}[ht]
\begin{center}
\centerline{\includegraphics[width=0.9\textwidth]{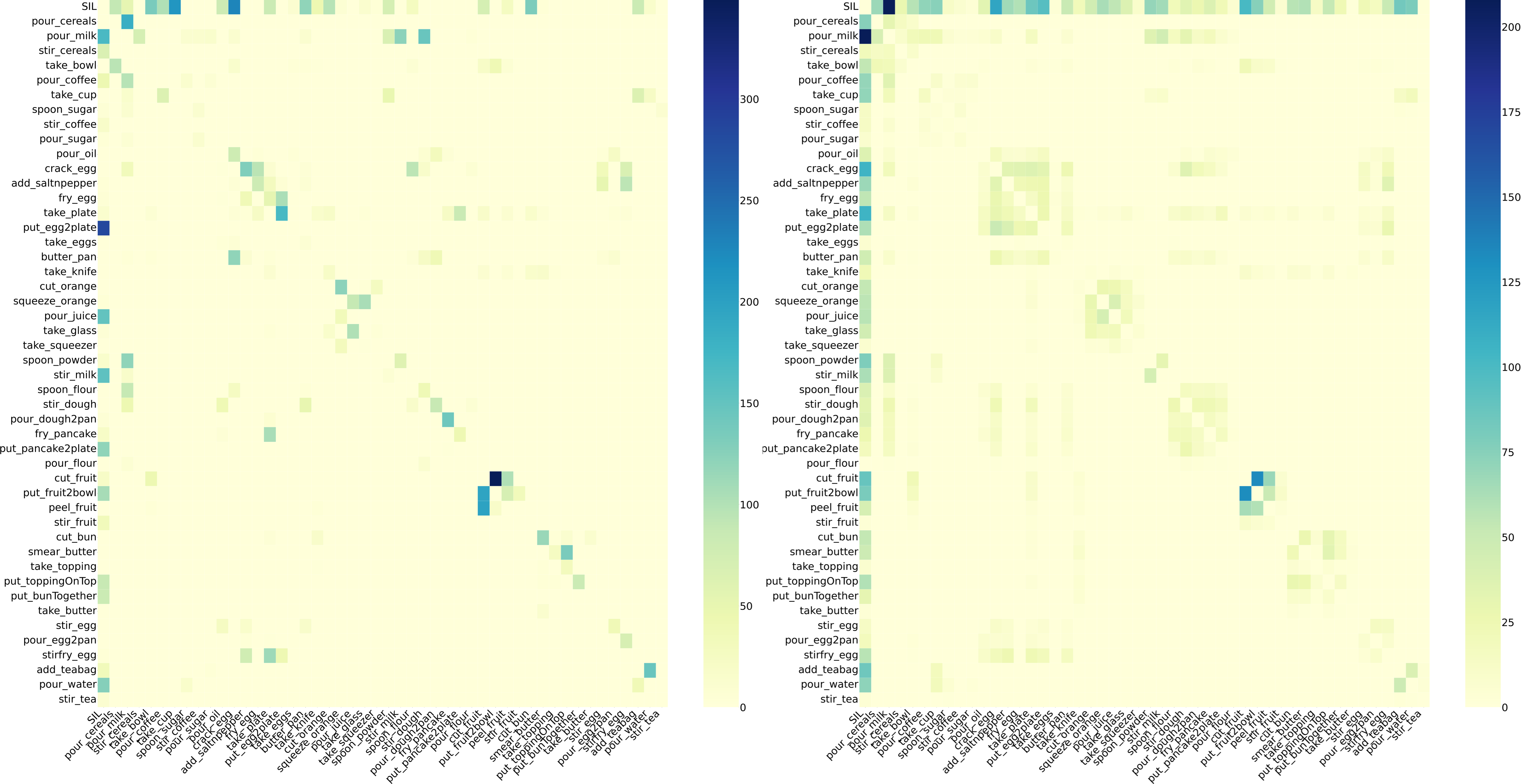}}
\caption{
\textbf{Heatmap of the frequency of 2-gram action label pairs in the Breakfast dataset.}
The left displays the original dataset and the right represents the dataset with the Shuffle technique.
The former action is represented on the Y-axis and the latter action on the X-axis.
}
\label{fig:app_heat_breakfast}
\end{center}
\end{figure}

\autoref{fig:app_heat_mask} illustrates the visualization of the frequency of the pair of action labels of 2 grams within the Breakfast dataset using the action masking method.
Furthermore, \autoref{fig:app_heat_50salads} presents results from the 50Salads dataset with the Shuffle Dataset approach, while \autoref{fig:app_heat_breakfast} shows results from the Breakfast dataset also employing the Shuffle Dataset technique.

\subsection{Qualitative results of Model  Prediction on Action Masking}
\label{subsec:qual_mask}
\begin{figure}[ht]
\begin{center}
\includegraphics[width=0.9\textwidth]{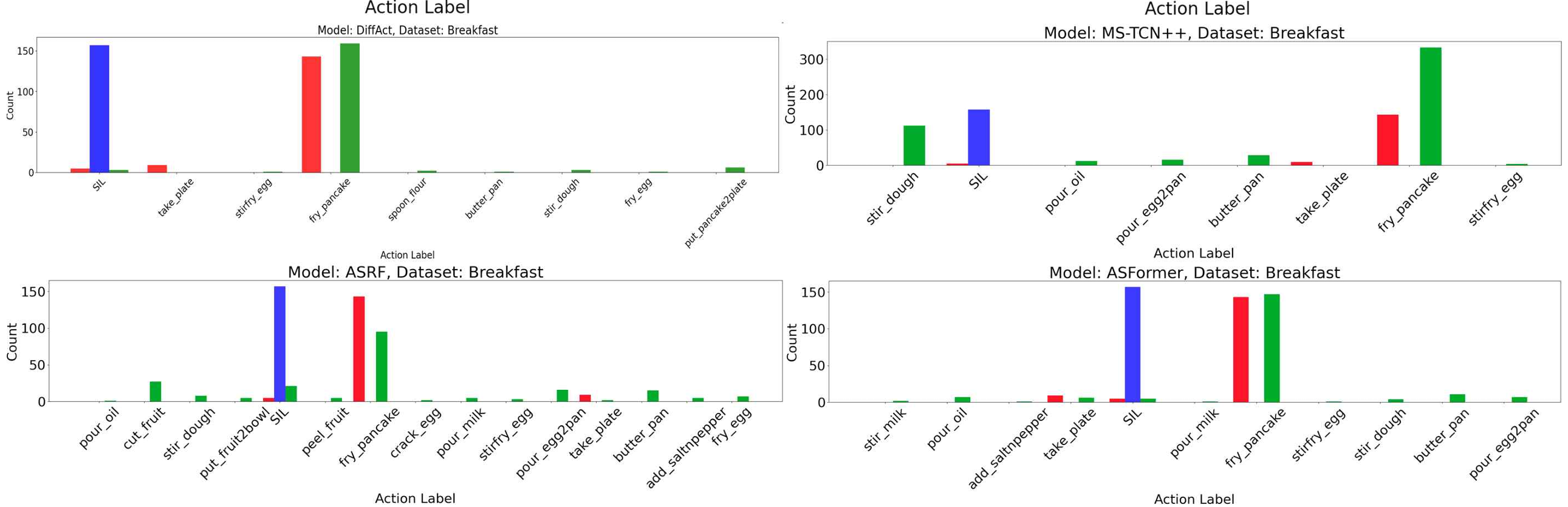}
\caption{
\textbf{Distribution of predicted action labels with four models on Breakfast dataset}
The \textcolor{red}{red bar} represents the count in the original video set;
the \textcolor{blue}{blue bar} displays the count of ground truth label in the masked video set used for evaluation, where the latter action label is replaced with `no-action' (`SIL' in Breakfast).
The \textcolor{green}{green bar} represents the count of the model's prediction for the masked video section.
}
\label{fig:app_mask}
\end{center}
\end{figure}

\autoref{fig:app_mask} shows the distribution of predicted and ground truth labels in breakfast with the applied action masking technique of the data set.
In this visualization, we selected `(pour\_dough2pan, fry\_pancake)' pair.

\subsection{Qualitative results of Model Prediction on Limited Shuffling}
\label{subsec:qual_shuf}
\begin{figure}[ht]
\begin{center}
\includegraphics[width=0.9\textwidth]{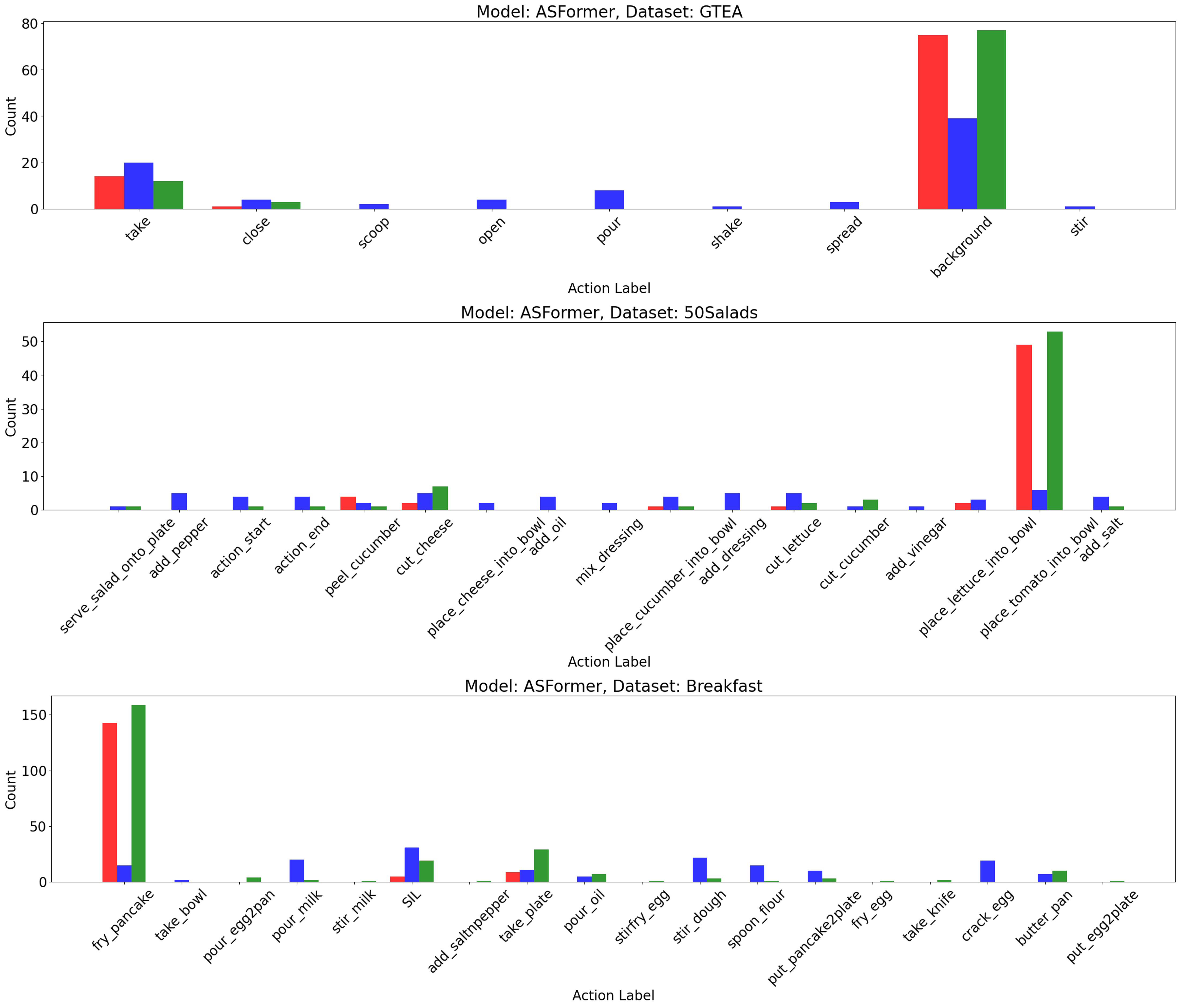}
\caption{
\textbf{Distribution of predicted action labels on ASFormer with various datasets.
}
}
\label{fig:var1}
\end{center}
\end{figure}

\begin{figure}[ht]
\begin{center}
\includegraphics[width=0.9\textwidth]{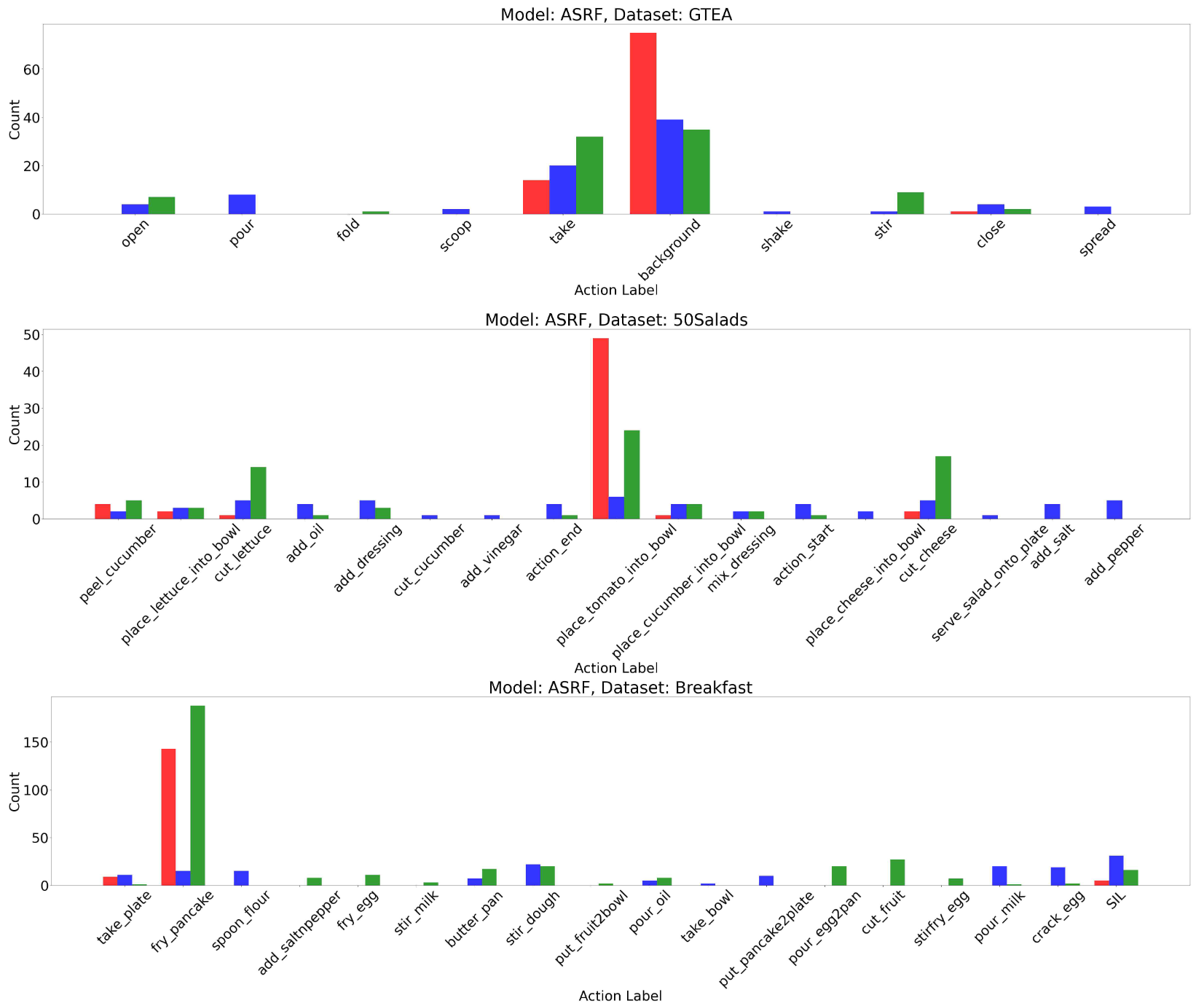}
\caption{
\textbf{Distribution of predicted action labels on ASRF with various datasets.
}
}
\label{fig:var2}
\end{center}
\end{figure}

\begin{figure}[ht]
\begin{center}
\includegraphics[width=0.9\textwidth]{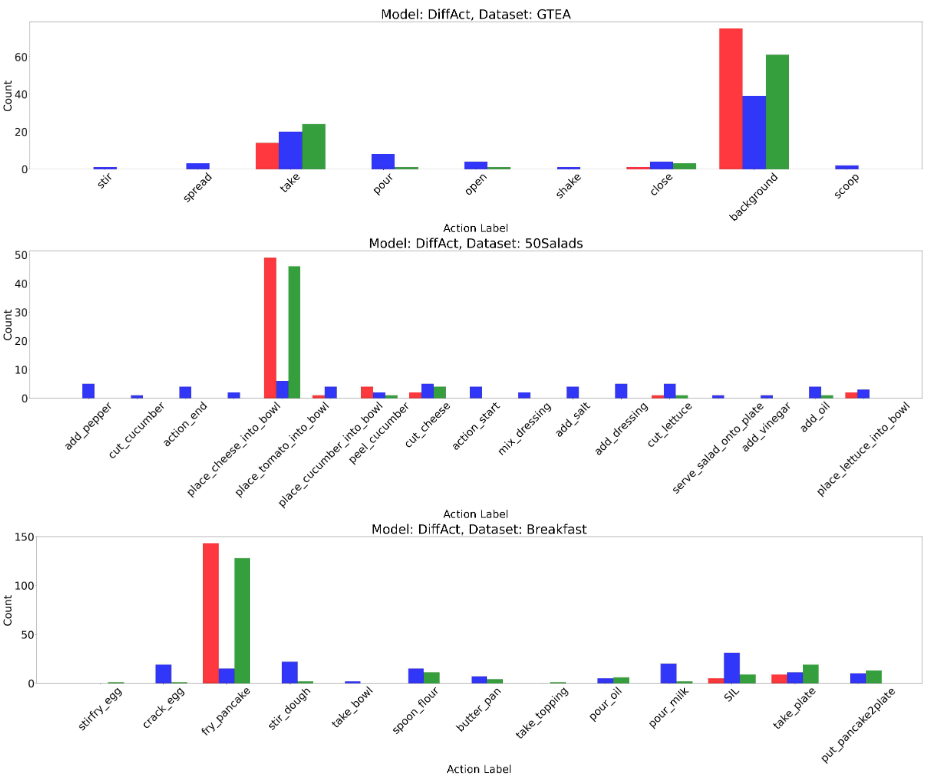}
\caption{
\textbf{Distribution of predicted action labels on DiffAct with various datasets.
}
}
\label{fig:var3}
\end{center}
\end{figure}

\begin{figure}[ht]
\begin{center}
\includegraphics[width=0.9\textwidth]{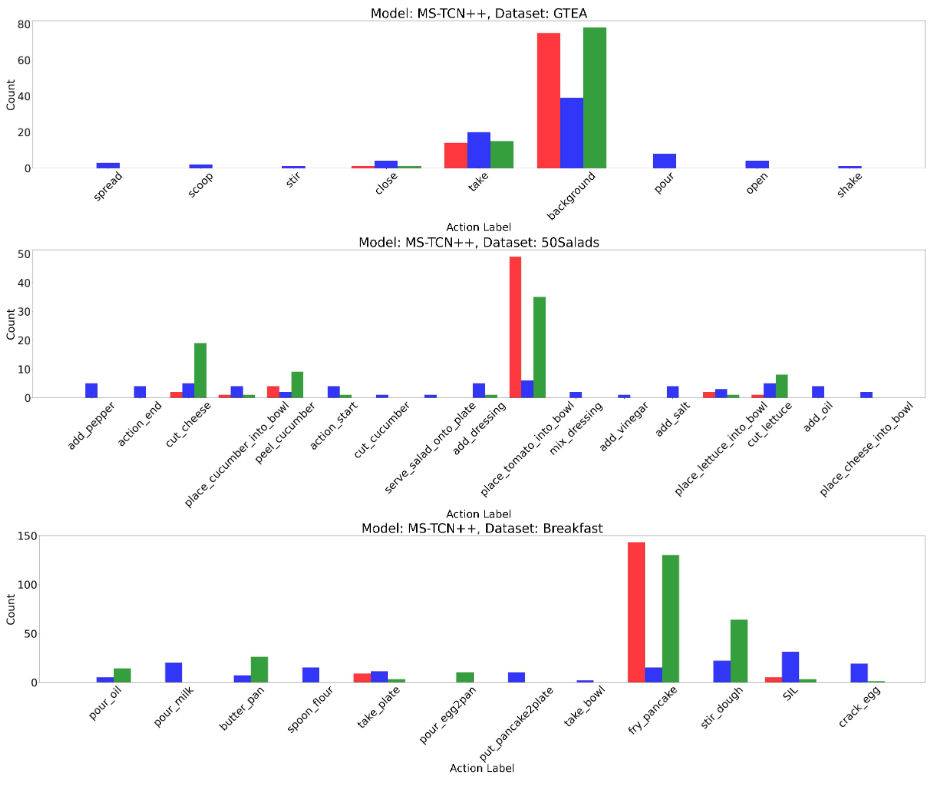}
\caption{
\textbf{Distribution of predicted action labels on MS-TCN++ with various datasets.
}
}
\label{fig:var4}
\end{center}
\end{figure}

\autoref{fig:var1}, \autoref{fig:var2}, \autoref{fig:var3}, and \autoref{fig:var4} show the distribution of the original label, limited shuffling label, and the prediction of the model in the limited shuffling dataset. The `(put, take)' action pair is selected for GTEA, `(cut\_tomato, place\_tomato\_into\_bowl)’ for 50Salads, and `(pour\_dough2pan, fry\_pancake)' for Breakfast, respectively.

\section{Positional Distibution of Action Labels in Dataset}
\begin{figure}[ht]
\begin{center}
\includegraphics[width=0.9\textwidth]{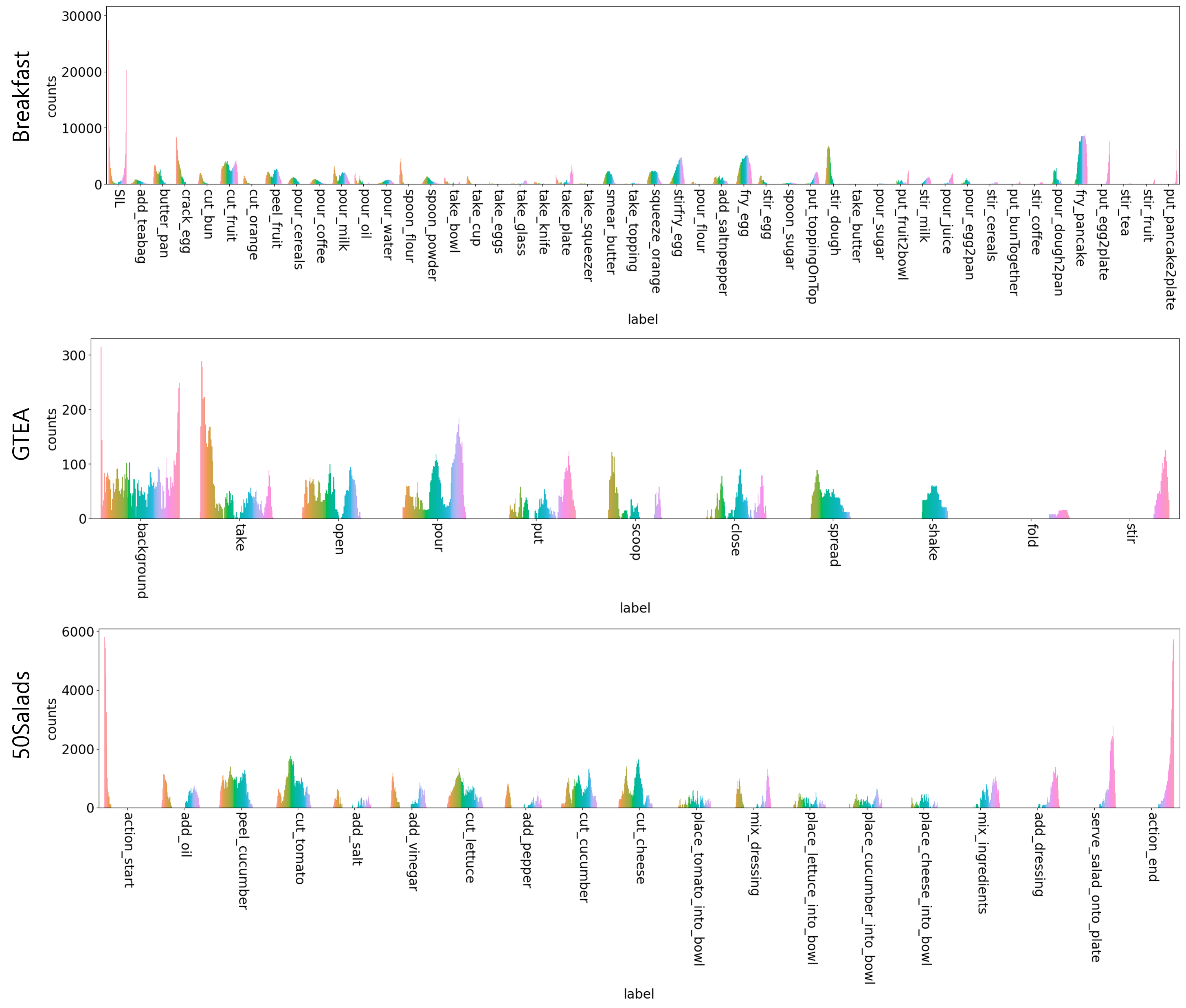}
\caption{
\textbf{Distribution of relative temporal position of action labels.}
Y-axis represent count, and X-axis is the normalized time (from 0 to 1) for each action label.
}
\label{fig:relative}
\end{center}
\end{figure}

In the context of comprehending instructional video learning, not only the interrelations among actions but also the temporal positional information of actions can help improve the accuracy of action prediction.
Nevertheless, akin to the concept of ordinal bias, this scenario may exemplify another form of bias.
\autoref{fig:relative} illustrates the normalized temporal positions corresponding to each label, highlighting the potential for models to be influenced by this information.

\end{document}